\documentclass[journal]{IEEEtran}

\usepackage{graphicx}
\usepackage{comment}
\usepackage{amsmath,amssymb} 
\usepackage{color}
\usepackage{caption}
\usepackage{subcaption}
\usepackage{boldline}
\usepackage{tabu}
\usepackage{multirow}
\usepackage{arydshln}
\usepackage{diagbox}
\usepackage{pifont}
\usepackage{threeparttable}

\newcommand{\eg}{\textit{e}.\textit{g}.}
\newcommand{\etal}{\textit{et al}.}
\newcommand{\ie}{\textit{i}.\textit{e}.,}

\begin{document}
%
\title{Interactive Knowledge Distillation for Image Classification}
%
%
%

\author{Shipeng~Fu,
        Zhen~Li,
        Jun~Xu,
        Mingming~Cheng,
        Zitao~Liu,
        and~Xiaomin~Yang
\thanks{S. Fu and X. Yang are with College of Electronics and Information Engineering, Sichuan University, Chengdu 610064, China. Email: S. Fu: fushipeng97@gmail.com, X. Yang: arielyang@scu.edu.cn}
\thanks{Z. Li, J. Xu and M. Cheng are with College of Computer Science, Nankai University, Tianjin 300350, China. Email: Z. Li: zhenli1031@gmail.com, J. Xu: nankaimathxujun@gmail.com, M. Cheng: cmm@nankai.edu.cn}
\thanks{Z. Liu is with TAL Education Group, Beijing 100080, China. Email: liuzitao@100tal.com}}

\markboth{Journal of \LaTeX\ Class Files,~Vol.~14, No.~8, August~2015}%
{Shell \MakeLowercase{\textit{et al.}}: Bare Demo of IEEEtran.cls for IEEE Journals}
%

\maketitle

\begin{abstract}
Knowledge distillation (KD) is a standard teacher-student learning framework to train a light-weight student network under the guidance of a well-trained, large teacher network.\ 
As an effective teaching strategy, interactive teaching has been widely employed at school to motivate students, in which teachers not only provide knowledge, but also give constructive feedback to students upon their responses, to improving their learning performance.\ 
In this work, we propose Interactive Knowledge Distillation (IAKD) to leverage the interactive teaching strategy for efficient knowledge distillation.\
In the distillation process, the interaction between the teacher network and the student one is implemented by swapping-in operation: randomly replacing the blocks in the student network with the corresponding blocks in the teacher network.\
In this way, we directly involve the teacher's powerful feature transformation ability for largely boosting the performance of the student network.\
Experiments with typical settings of teacher-student networks demonstrate that the student networks trained by our IAKD achieve better performance than those trained by conventional knowledge distillation methods on diverse image classification datasets.
\end{abstract}


%
\IEEEpeerreviewmaketitle

\section{Introduction}
\label{sec:intro}
\IEEEPARstart{O}{ver} the past few years, deeper and deeper convolutional neural networks (CNNs) have demonstrated cutting edge performance on various computer vision tasks~\cite{huang2017densely,karras2019style,SunXLW19,li2019srfbn,gao2019res2net}.\ 
However, they usually undergo huge computational costs and memory consumption, and are difficult to be embedded into resource-constrained devices, \eg, mobiles and UAVs.\ 
To reduce the resource consumption while maintaining good performance, researchers leverage knowledge distillation techniques~\cite{luo2016face,chen2017learning,chen2018darkrank,wang2019distilling,liu2019structured,hou2019learning} to transfer informative knowledge from a cumbersome but well-trained teacher network into a light-weight but unskilled student network.\
\begin{figure}[htbp]
\centering
\includegraphics[width=0.47\textwidth]
{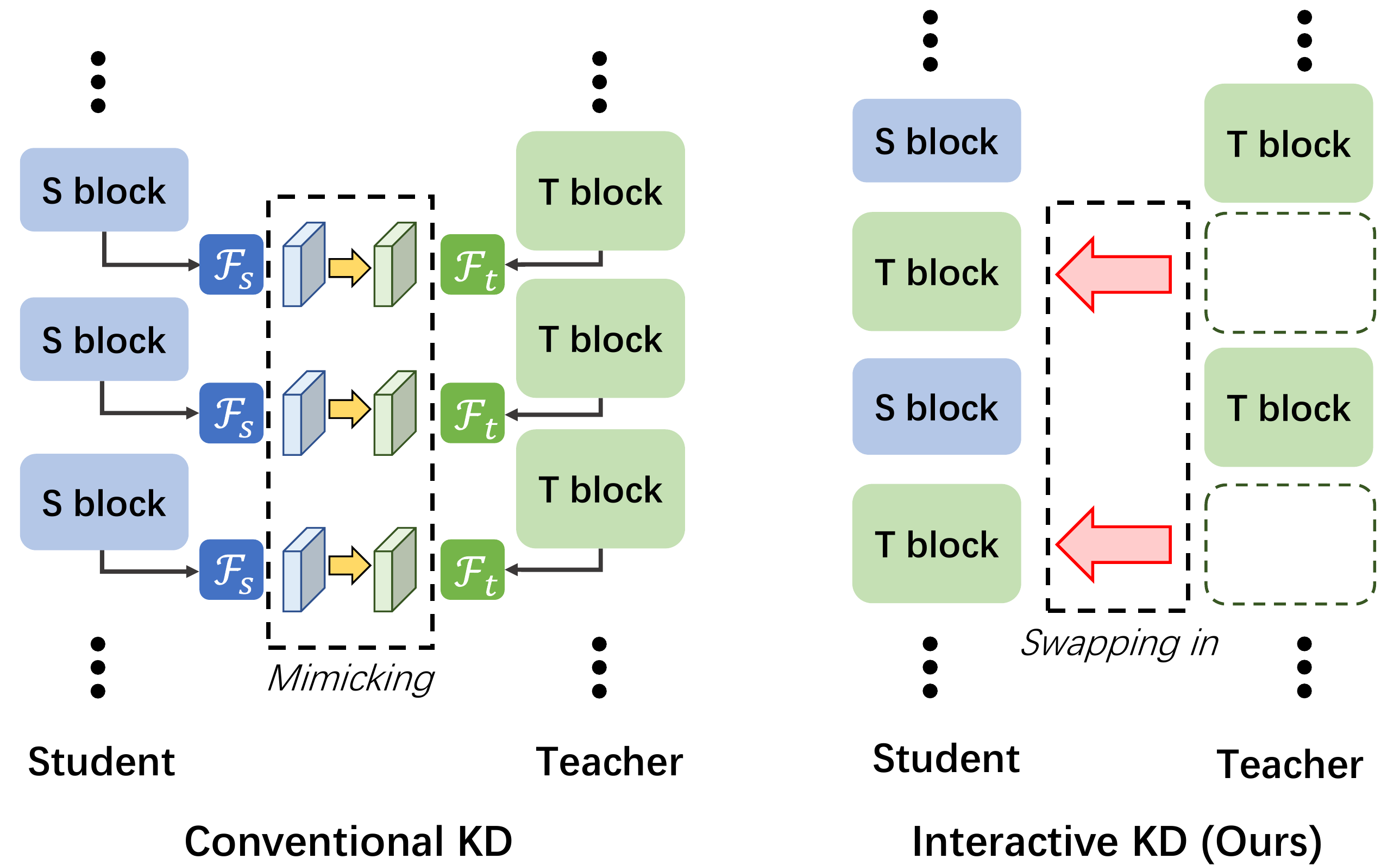}
\caption{\textbf{Differences between our proposed interactive knowledge distillation method and conventional, non-interactive ones}.\ ``S block" denotes the student block, ``T block" denotes the teacher block. $\mathcal{F}_{s}$ and $\mathcal{F}_{t}$ denote the definitions of the student and the teacher knowledge, respectively, and vary a lot from method to method.}
\label{fig:diff_distill}
\end{figure} 

Most of existing knowledge distillation methods~\cite{hinton2015distilling,Zagoruyko2017AT,park2019relational,heo2019overhaul,tian2019crd,mirzadeh2019improved} encourage the student network to mimic the representation space of the teacher network for approaching its proficient performance.\ 
Those methods manually define various kinds of knowledge based on the teacher network responses, such as softened outputs~\cite{hinton2015distilling}, attention maps~\cite{Zagoruyko2017AT}, flow of solution procedure~\cite{yim2017gift}, and relations~\cite{park2019relational}.\ These different kinds of knowledge facilitate the distillation process through additional distillation losses.\ 
If we regard the responses of the teacher network as the theorems in a textbook written by the teacher, manually defining those different kinds of knowledge just rephrases those theorems and further points out an easier way to make the student understand those theorems better. 
This kind of distillation process is viewed as \textit{non-interactive} knowledge distillation, since the teacher network just sets a goal for the student network to mimic, ignoring the interaction with the student network.\

The \textit{non-interactive} knowledge distillation methods undergo one major problem: since the feature transformation ability of the student network is less powerful than that of the teacher network, the knowledge that the student has learned through mimicking is impossible to be identical with the knowledge provided by the teacher, which may impede the knowledge distillation performance.\ 
This problem becomes more tough when introducing multi-connection knowledge~\cite{Zagoruyko2017AT,ABdistill}.\ 
Specifically, since the student can not perfectly mimic the teacher's knowledge even if in the shallow blocks, the imperfectly-mimicked knowledge is reused in the following blocks to imitating the corresponding teacher's knowledge. 
Therefore, the gap between the knowledge learned by the student and the one provided by the teacher becomes larger, restricting the performance of knowledge distillation. \ 

Based on the discussion above, we try to make the distillation process \textit{interactive} and the teacher network truly involved in guiding the student network. 
Specifically, the student first takes actions according to the problems he faces, then the teacher gives feedback based on the student's actions, finally the student takes actions according to the teacher's feedback.
As shown in Fig.~\ref{fig:diff_distill}, instead of forcing the student network to mimic the teacher's representation space, our interactive distillation directly involves the teacher's powerful feature transformation ability into the student network.\ 
Consequently, the student can make the best use of the improved features to better exploit its potential on relevant tasks.

In this paper, we propose a simple yet effective knowledge distillation method named Interactive Knowledge Distillation (IAKD).\ 
In IAKD, we randomly swap in the blocks in the teacher network to replacing the blocks in the student network during the distillation phase.\
Each set of swapped-in teacher blocks responds to the output of previous student block, and provides better feature maps to motivate the next student block. 
In addition, randomly swapping in the teacher blocks makes it possible for the teacher to guide the student in many different manners. 
Compared with other conventional knowledge distillation methods, our proposed method does not deliberately force the student's knowledge to be similar to the teacher's knowledge. 
Therefore, we do not need the additional distillation losses to drive the knowledge distillation process, resulting in no need of hyper-parameters to balancing the task-specific loss and distillation losses.
Besides, the distillation process of our IAKD method is highly efficient.\ The reason is that our IAKD discards the cumbersome knowledge transformation process and gets rid of the feature extraction of the teacher network, both of which commonly exist in conventional knowledge distillation methods.\ 
Experimental results demonstrate that our proposed method effectively boosts the performance of the student, which proves that the student can actually benefit from the interaction with the teacher.

\section{Related Work}
Since Ba and Caruana~\cite{ba2014deep} first introduced the view of teacher supervising student for model compression, many variant works have emerged based on their teacher-student learning framework. 
To achieve promising model compression results, these works focus on how to better capture the knowledge of the large teacher network to supervise the training process of the small student network. 
The most representative work about knowledge distillation (KD) comes from Hinton~\etal~\cite{hinton2015distilling}. 
They defined knowledge as the teacher's softened logits and encouraged the student to mimic it instead of the raw activations before softmax~\cite{ba2014deep}, because they argued that the intra-class and inter-class relationships learned by the pre-trained teacher network can be described more accurately by the softened logits than by the raw activations. 
However, simply using logits as the learning goal for the student may limit the information that the teacher distills to the student. 
To introduce more knowledge of the teacher network, FitNets~\cite{romero2014fitnets} added element-wise supervision on the feature maps at intermediate layers to assisting the training of the student. 
However, this approach only works well when one supervision loss is applied to one intermediate layer. It can not achieve satisfying results when more losses are added to supervising more intermediate layers~\cite{yim2017gift}.
This is because supervision on the feature maps at many different layers makes the constraints become too restrictive. 
There are many methods trying to soften the constraints while preserving meaningful information of feature maps.
Zagoruyko and Komodakis~\cite{krizhevsky2009learning} condensed feature maps to attention maps based on channel statistics of each spatial location.
Instead of defining knowledge based on feature maps of a single layer, Yim~\etal~\cite{yim2017gift} employed Gramian matrix to measure the correlations between feature maps from different layers. 
Besides, Kim~\etal~\cite{kim2018paraphrasing} utilized convolutional auto-encoder, and Lee~\etal~\cite{hyun2018self} used Singular Value Decomposition (SVD) to perform dimension reduction on feature maps. 
Heo~\etal~\cite{ABdistill,heo2019overhaul} demonstrated that simply using activation status of neurons could also assist the distillation process. 
Ahn~\etal~\cite{ahn2019variational} proposed a method based on information theory which maximizes the mutual information between the teacher network and the student network. 
To enable the student to acquire more meaning information, a lot of studies defined knowledge based on relations rather than individual feature maps~\cite{park2019relational,liu2019knowledge,lee2019graph,tung2019similarity}. 

On the other hand, our proposed IAKD is essentially different from aforementioned knowledge distillation methods.
IAKD does not require distillation losses to drive the distillation process (see Eq.~\ref{eq:ckd_loss} and Eq.~\ref{eq:hybrid_loss}). In the learning process of the student, we directly involve the teacher's powerful feature transformation ability to improve the relatively weak features extracted by the student, since better features are critical for getting a better prediction.
\section{Proposed Method}
\subsection{Conventional Knowledge Distillation}
\label{sec:ckd}
In general, the conventional, feature-based knowledge distillation methods~\cite{ABdistill,ahn2019variational,Zagoruyko2017AT,yim2017gift,heo2019overhaul} decompose the student network and the teacher one into $M$ modules. 
The student network is trained to minimize the following loss function:
\begin{equation}
	\label{eq:ckd_loss}
    \mathcal{L}_{CKD} =\gamma(y,\hat{y})+ \sum_{m \in M}\lambda_{m}\varphi_{m}(F_{m}^{S}, F_{m}^{T}),
\end{equation}
where $\gamma(\cdot,\cdot)$ is the task-specific loss.
$y$ represents the ground truth. 
$\hat{y}$ is the predicted result of the student network. 
$F_{m}^{S}$ and $F_{m}^{T}$ denote the transformed output of the student network and the one of the teacher network produced by the $m$-th module.
$\lambda_{m}$ is a tunable balance factor to balancing the different losses.
$\varphi_{m} (\cdot, \cdot)$ is the $m$-th distillation loss to narrowing down the difference between $F_{m}^{S}$ and $F_{m}^{T}$. 
$\varphi_{m} (\cdot, \cdot)$ varies from method to method because of different definitions of knowledge. 
As can be seen in Eq.~\ref{eq:ckd_loss}, the conventional knowledge distillation methods essentially aim at forcing the student to mimic the teacher's representation space.

\begin{figure}[t]
\centering
\includegraphics[width=.47\textwidth]{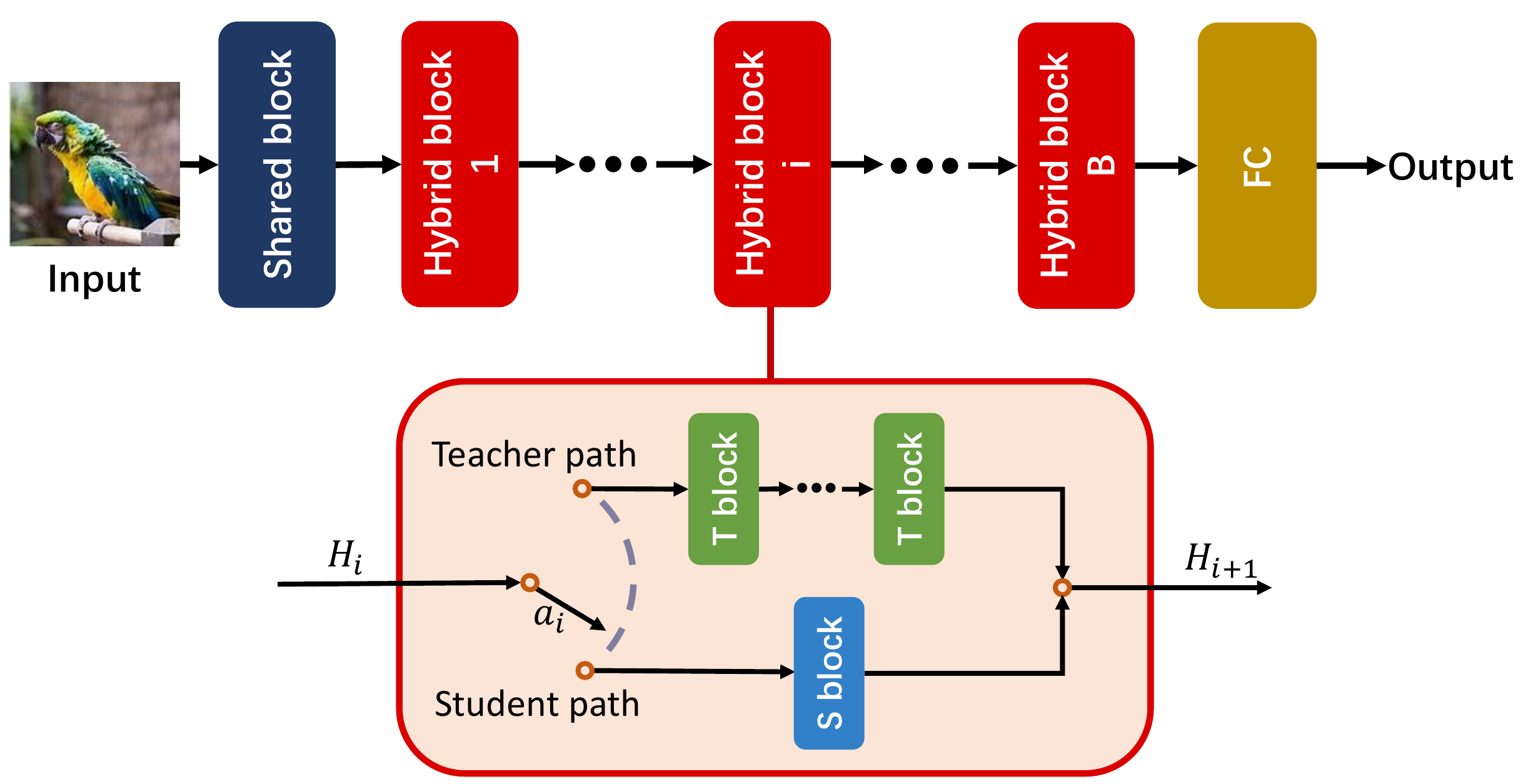}
\caption{The diagram of the hybrid network which can effectively implement the proposed interactive knowledge distillation (IAKD). Specifically, the hybrid network enables the teacher blocks to randomly replace the student block of the original student network. FC denotes the fully-connected layer.}
\label{fig:hybrid block}
\end{figure}

\subsection{Interactive Knowledge Distillation}
To achieve the interaction between the student network and the teacher network, our IAKD randomly swaps in the teacher blocks to replace the student blocks at each iteration.
The swapped-in teacher blocks can respond to the output of the previous student block and then provide better features to motivate the next student block.
To make all student blocks fully participate in the distillation process, after each training iteration, we put back the replaced student blocks to their original positions in the student network.\ Due to the return of replaced student blocks, the architecture of the student is kept consistent for the next swapping-in operation.

\noindent
\paragraph{Hybrid block for swapping-in operation}
%
As shown in Fig.~\ref{fig:hybrid block}, for effective implementation of such interactive strategy, we devise a two-path hybrid block, which is composed of one student block (the student path) and several teacher blocks to be swapped in (the teacher path).
We use the hybrid block to replace the original student block in the student network to form a hybrid network. 
However, the blocks and layers that are shared by the teacher and the student are not replaced by the hybrid blocks\footnote{The fully-connected layer and the transition blocks in ResNet~\cite{he2016deep} are considered as the shared parts.}, since the shared parts have the same learning capacity. 
The number of hybrid blocks depends on the number of non-shared student blocks.
Every non-shared student block will pair with one or several non-shared teacher blocks to form the hybrid block. 
All non-shared teacher blocks will be taken to form the hybrid blocks.

For the $i$-th hybrid block in the hybrid network, let $a_{i} \in \{0,1\}$ denotes a  Bernoulli random variable, which indicates whether the student path is selected ($a_{i}=1$) or the teacher path is chosen ($a_{i}=0$). 
Therefore, the output of the $i$-th hybrid block can be formulated as 
\begin{equation}
	\label{eq:i-th hybrid block}
	H_{i+1}=a_{i} f_{i}^{S}(H_{i})+(1-a_{i})f_{i}^{T}(H_{i}),
\end{equation}
where $H_{i}$ and $H_{i+1}$ denote the input and the output of the $i$-th hybrid block. $f_{i}^{S}(\cdot)$ is the function of the student block. $f_{i}^{T}(\cdot)$ is the nested functions of the teacher blocks. 
 
If $a_{i}=1$, the student path is chosen and Eq.~\ref{eq:i-th hybrid block} reduces to:
\begin{equation}
	\label{eq:i-th hybrid block ai=1}
	H_{i+1}=f_{i}^{S} (H_{i}).
\end{equation}
This means the hybrid block degrades to the original student block. 
If $a_{i}=0$, the teacher path is chosen and Eq.~\ref{eq:i-th hybrid block} reduces to: 
\begin{equation}
	\label{eq:i-th hybrid block ai=0}
	H_{i+1}=f_{i}^{T}(H_{i}).
\end{equation}
This implies the hybrid block degrades to the teacher blocks. 
Next, we denote the probability of selecting the student path in the $i$-th hybrid block as $p_{i}=P(a_{i}=1)$. 
Therefore, each student block is randomly replaced by the teacher blocks with probability $(1-p_{i})$. 
The swapped-in teacher blocks can interact with the student blocks to inspire the potential of the student.

In addition, the student blocks are replaced by the teacher blocks in a random manner, which means the teacher can guide the student in many different situations through interaction. 
In contrast, swapping in the teacher blocks by manually-defined manners allows the teacher to guide the student only in limited situations. 
%
%
Therefore, to traverse as many different situations as possible, the student blocks are replaced by the teacher blocks in a random manner.

\paragraph{Distillation phase}
%
When the image features go through each hybrid block, the path is randomly chosen during the distillation phase. Let us consider two extreme situations. 
The first is that $p_{i}=1$ for all hybrid blocks. 
The student path is chosen in every hybrid block and the hybrid network becomes the student network.
The second is that $p_{i}=0$ for all hybrid blocks.
Under such circumstances, all teacher paths are chosen, and the hybrid network becomes the teacher network. 
In other cases ($0<p_{i}<1$), the input randomly goes through the student path or the teacher path in each hybrid block.
The swapped-in teacher blocks interact with the student blocks, thus they can respond to the output of the previous student block and further provide better features to motivate the following student block.
Before distillation phase, the parameters of the teacher blocks in each hybrid block are loaded from the corresponding teacher blocks in the pretrained teacher network. 
Next, during distillation process, those parameters are frozen, and the batch normalization layers still use the mini-batch statistics. 
The student block in each hybrid block, and other shared parts are randomly initialized. 
The optimized loss function of the hybrid network $\mathcal{L}_{h}$ can be formulated as
\begin{equation}
	\label{eq:hybrid_loss}
    \mathcal{L}_{h} =\gamma(y,\hat{y}_{h}),
\end{equation}
where $\gamma(\cdot,\cdot)$ is the task-specific loss. $y$ denotes the ground truth. $\hat{y}_{h}$ is the predicted result of the hybrid network. 
At each iteration, only the student parts which the feature maps go through are updated (the shared parts are considered as the student parts since they are trained from scratch).

Comparing Eq.~\ref{eq:hybrid_loss} with Eq.~\ref{eq:ckd_loss}, we can see our proposed method does not need the additional distillation losses to drive the knowledge distillation process.
Therefore, laboriously searching the optimal hyper-parameters to balance the task-specific loss and different distillation losses can be avoided. 
Besides, the complicated knowledge transformation and extra forward computation of the teacher network are unnecessarily required by our proposed method, leading to an efficient distillation process (see Sec.~\ref{sec:4.3} for training time comparison).

\paragraph{Test phase} 
Since knowledge distillation aims at improving the performance of the student network, we just test the whole student network to check whether the student can achieve performance improvement or not. 
For convenient validation, we fix the probability $p_{i}$ to 1 in the test phase. 
Under this circumstances, the hybrid network becomes the student network during the test phase. 
Meanwhile, for practical deployment, it is not reasonable to directly use the cumbersome hybrid network. We can reduce the burden of resource consumption by separating the student network from the hybrid network for further inference. 
\subsection{Probability Schedule}
\label{sec:probabiltiy schedule}
The interaction level between the teacher network and the student network is controlled by $p_{i}$. 
The smaller value of $p_{i}$ indicates the student network has more interaction with the corresponding teacher blocks. 
For simplicity, we assume $p_{1}=p_{2}=\dots = p_{i} = \dots = p_{B}$, which suggests that every hybrid block shares the same interaction rule.
To properly utilize the interaction mechanism, we propose three kinds of probability schedules for the change of $p_{i}$: 1) uniform schedule, 2) linear growth schedule and 3) review schedule. These schedules are discussed in detail as follows. 

\begin{figure}
     \centering
     \begin{subfigure}[b]{0.235\textwidth}
         \centering
         \includegraphics[width=\textwidth]{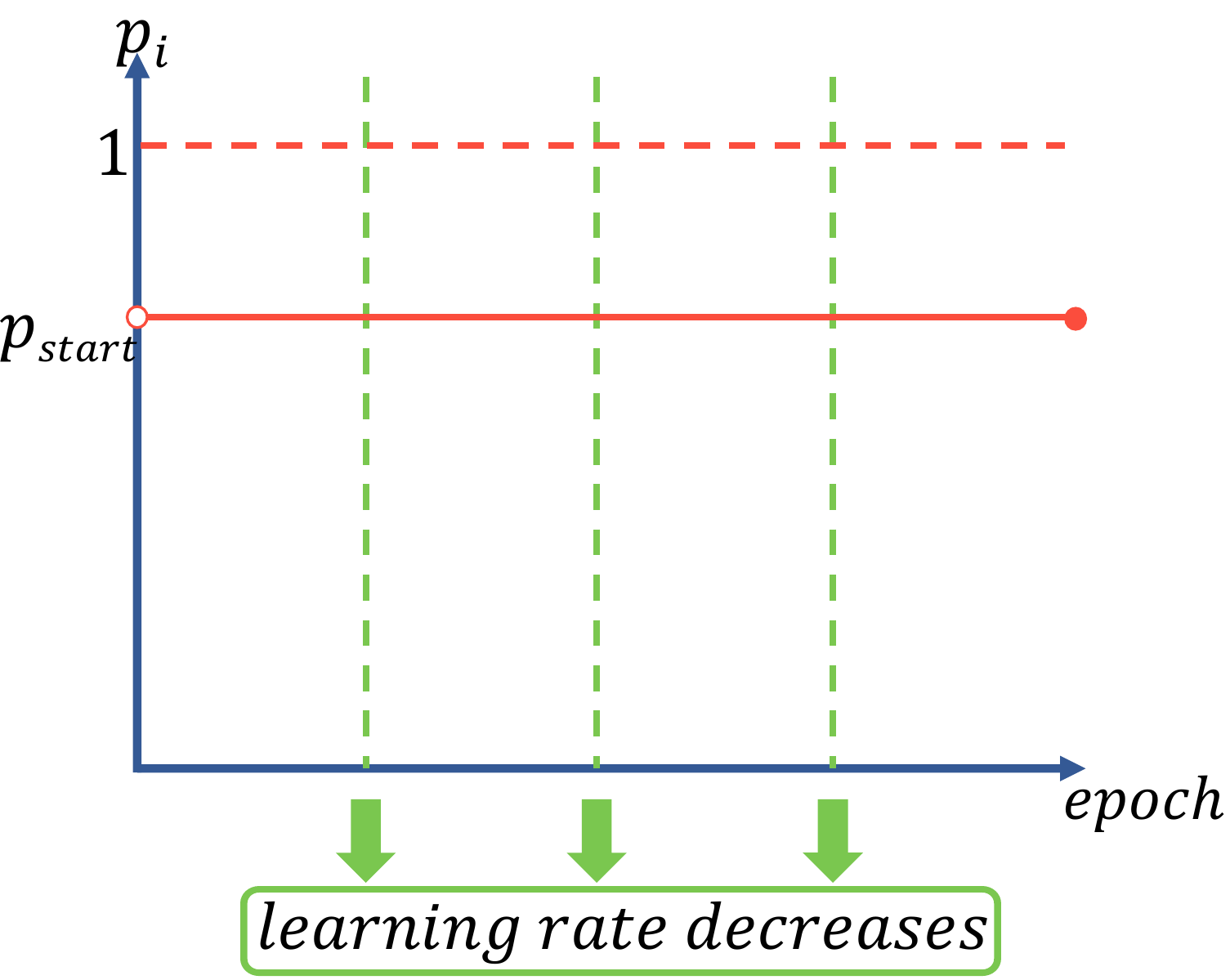}
         \caption{Uniform schedule}
         \label{fig:uniform schedule}
     \end{subfigure}
     \hfill
     \begin{subfigure}[b]{0.235\textwidth}
         \centering
         \includegraphics[width=\textwidth]{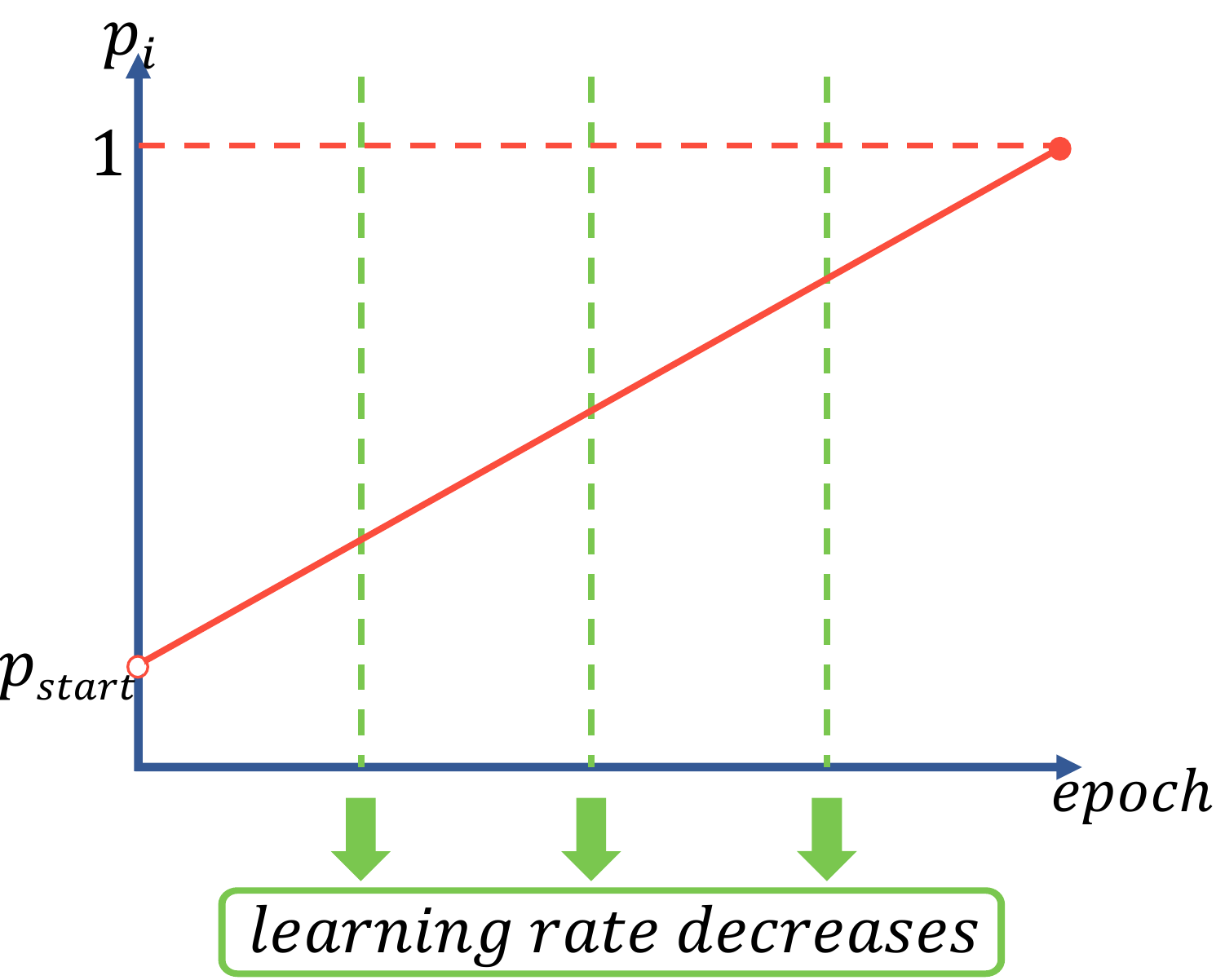}
         \caption{Linear growth schedule}
         \label{fig:linear growth schedule}
     \end{subfigure}
     \hfill
     \begin{subfigure}[b]{0.235\textwidth}
         \centering
         \includegraphics[width=\textwidth]{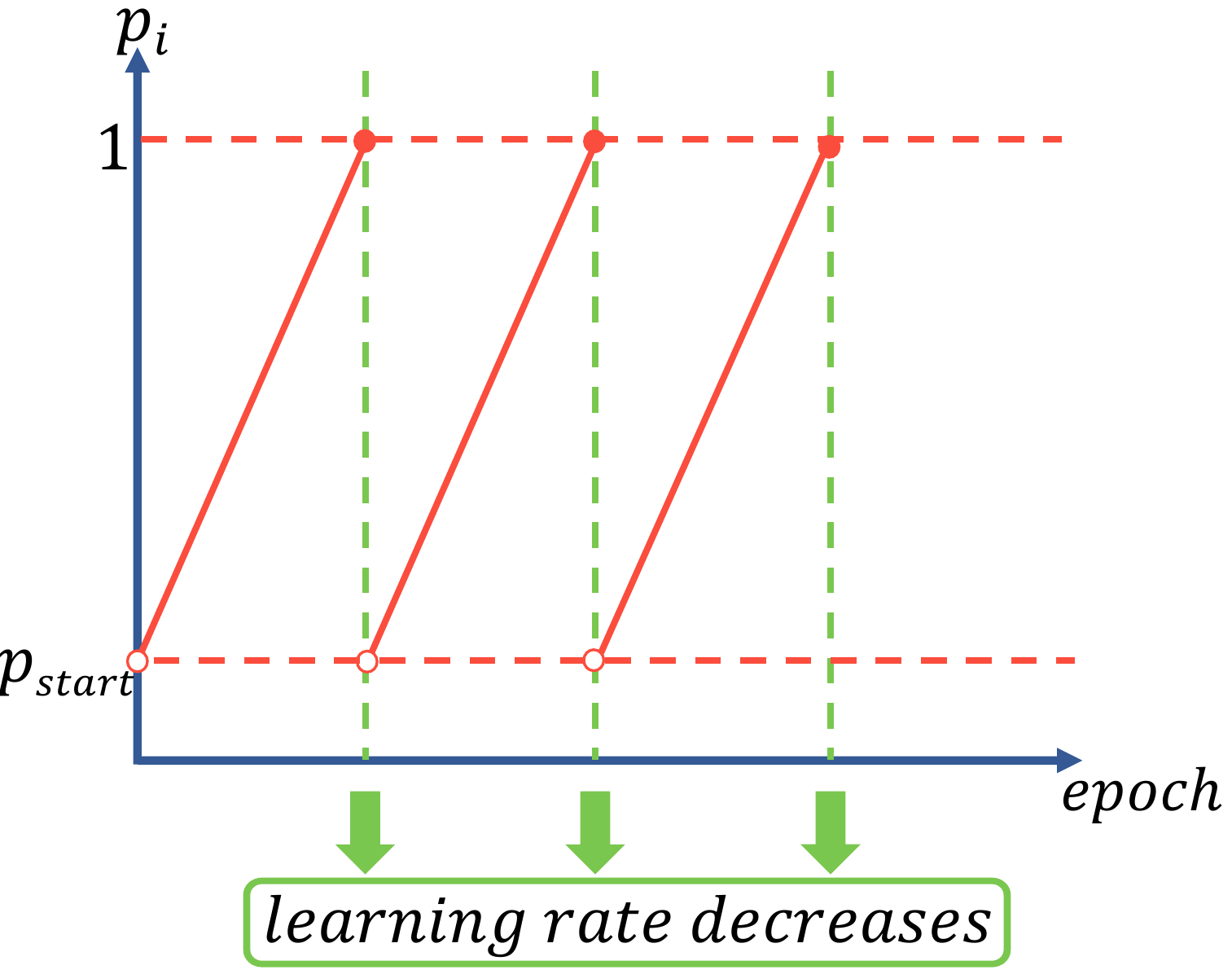}
         \caption{Review schedule}
         \label{fig:review schedule}
     \end{subfigure}
        \caption{Schematic illustrations for $p_{i}$ with different probability schedules. The vertical green dashed lines indicate when the learning rate decreases.}
        \label{fig:different schedules}
\end{figure}

\paragraph{Uniform schedule}
The simplest probability schedule is to fix $p_{i}=p_{start}\in(0,1)$ for all epochs during the distillation phase. 
This means the interaction level does not change. 
See Fig.~\ref{fig:uniform schedule} for a schematic illustration. 

\paragraph{Linear growth schedule}
With the epoch increasing, the teacher network could participate less in the distillation process as \cite{romero2014fitnets,yim2017gift,ABdistill} do, because the student network gradually becomes more powerful to handle the task by itself.
Thus, we set the value of $p_{i}$ according to a function of the number of epoch, and propose a simple linear growth schedule. 
The probability of choosing the student path in each hybrid block increases linearly from $p_{i}=p_{start}\in(0,1)$ for the first epoch to $p_{i}=1$ for the last epoch (see  Fig.~\ref{fig:linear growth schedule}). 
The linear growth schedule indicates that as the knowledge distillation process proceeds, the interaction level between the student and the teacher gradually decreases. 
Vividly speaking, the teacher teaches the student through frequent interaction in early epochs, and the student attempts to solve the problem by himself in late epochs.

\paragraph{Review schedule}
The proposed schedules above ignore the change of the learning rate during training. When the learning rate decreases, the optimization algorithm narrows the parameter search radius for more refined adjustments. 
Thus, we reset $p_{i}$ to $p_{start}$ when the learning rate decreases so that the teacher can \textit{re-interact} with the student for better guidance. 
Within the interval where the learning rate remains unchanged, the probability increases linearly from $p_{i}=p_{start}\in(0,1)$ to  $p_{i}=1$. 
Fig.~\ref{fig:review schedule} shows a schematic illustration. 
We call this kind of probability schedule as the review schedule because it is like the teacher gives review lessons after finishing a unit so that the student can grasp the knowledge more firmly. 

The effectiveness of different probability schedules will be validated experimentally in Sec.~\ref{sec.4.1}.
\section{Experiments}
\label{sec:experiments}
In this section, we first investigate the effectiveness of different probability schedules and then conduct a series of experiments to analyze the impact of the interactive mechanism on the distillation process. 
At last, we compare our proposed IAKD with conventional, \textit{non-interactive} methods on several public classification datasets to demonstrate the superiority of our proposed method. 
\subsection{Different Probability Schedules}
\label{sec.4.1}
To investigate the effectiveness of different probability schedules, we conduct experiments on CIFAR-10~\cite{krizhevsky2009learning} and CIFAR-100~\cite{krizhevsky2009learning} datasets. 
The CIFAR-10 dataset consists of 60000 images of size $32\times32$ in 10 classes. 
The CIFAR-100 dataset~\cite{krizhevsky2009learning} has 60000 images of size $32\times32$ from 100 classes.
For all experiments in this subsection, we use SGD optimizer with momentum 0.9 to optimize the hybrid network. 
We set batchsize to 128 and set weight decay to $1\times10^{-4}$.
The hybrid network is trained for 200 epochs. 
The learning rate starts at 0.1 and is multiplied by 0.1 at 100, 150 epochs. 
Images of size $32\times32$ are randomly cropped from zero-padded $40\times40$ images. 
Each image is horizontally flipped with a probability of 0.5 for data augmentation.
\subsubsection{Uniform Schedule}
We first conduct experiments with the uniform schedule mentioned in Sec.~\ref{sec:probabiltiy schedule}. 
We employ ResNet~\cite{he2016deep} as our base network architecture and denote ResNet-d as the ResNet with a specific depth d. 
ResNet-44 and ResNet-26 are used as the teacher/student (T/S) pair. 
Therefore, in each hybrid block, there is one student block in the student path, and two teacher blocks in the teacher path. 

\begin{figure}
     \centering
     \begin{subfigure}[b]{0.24\textwidth}
         \centering
         \includegraphics[width=\textwidth]{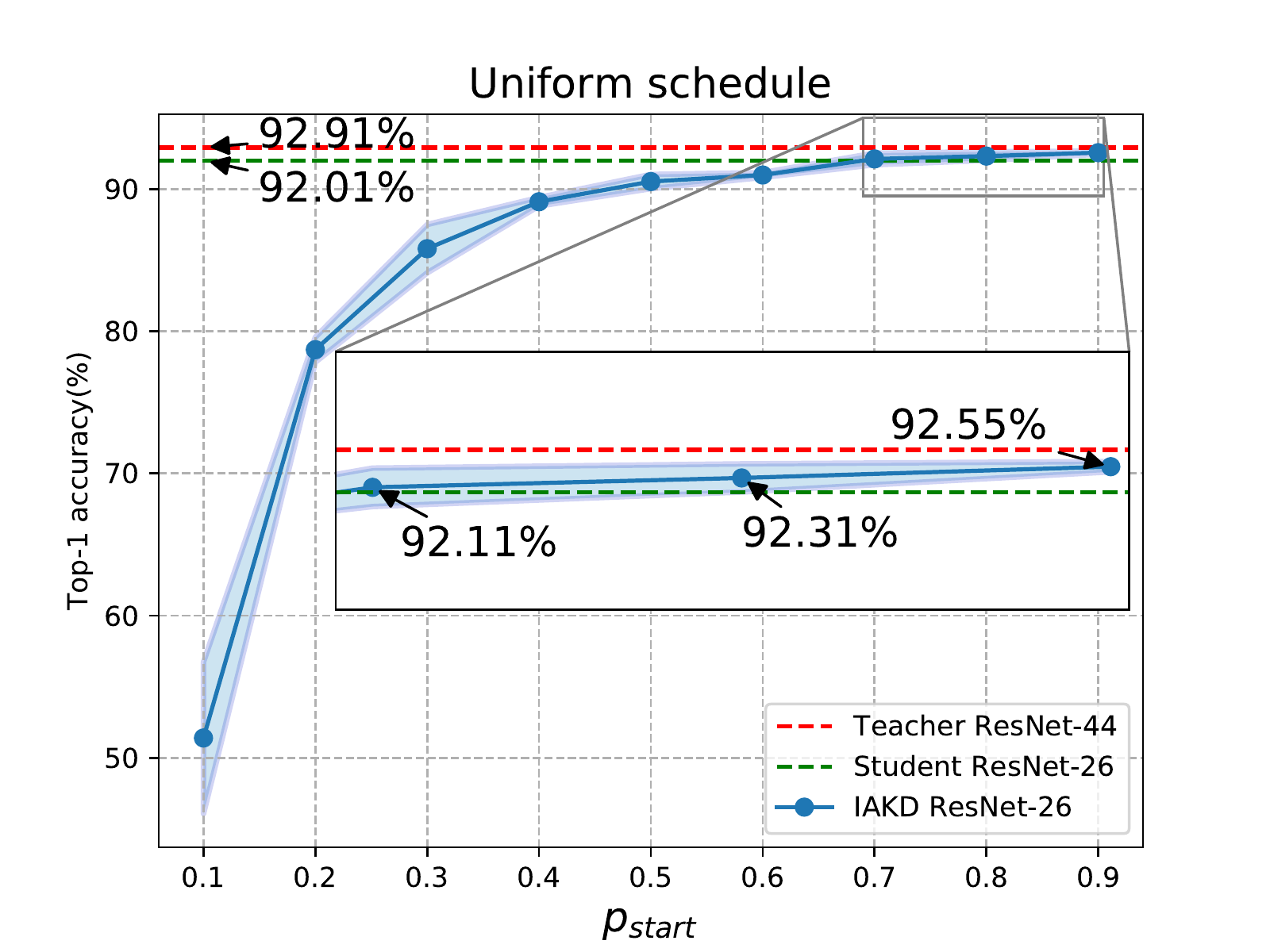}
         \caption{CIFAR-10}
         \label{fig:cifar10_uniform_results}
     \end{subfigure}  
     \begin{subfigure}[b]{0.24\textwidth}
         \centering
         \includegraphics[width=\textwidth]{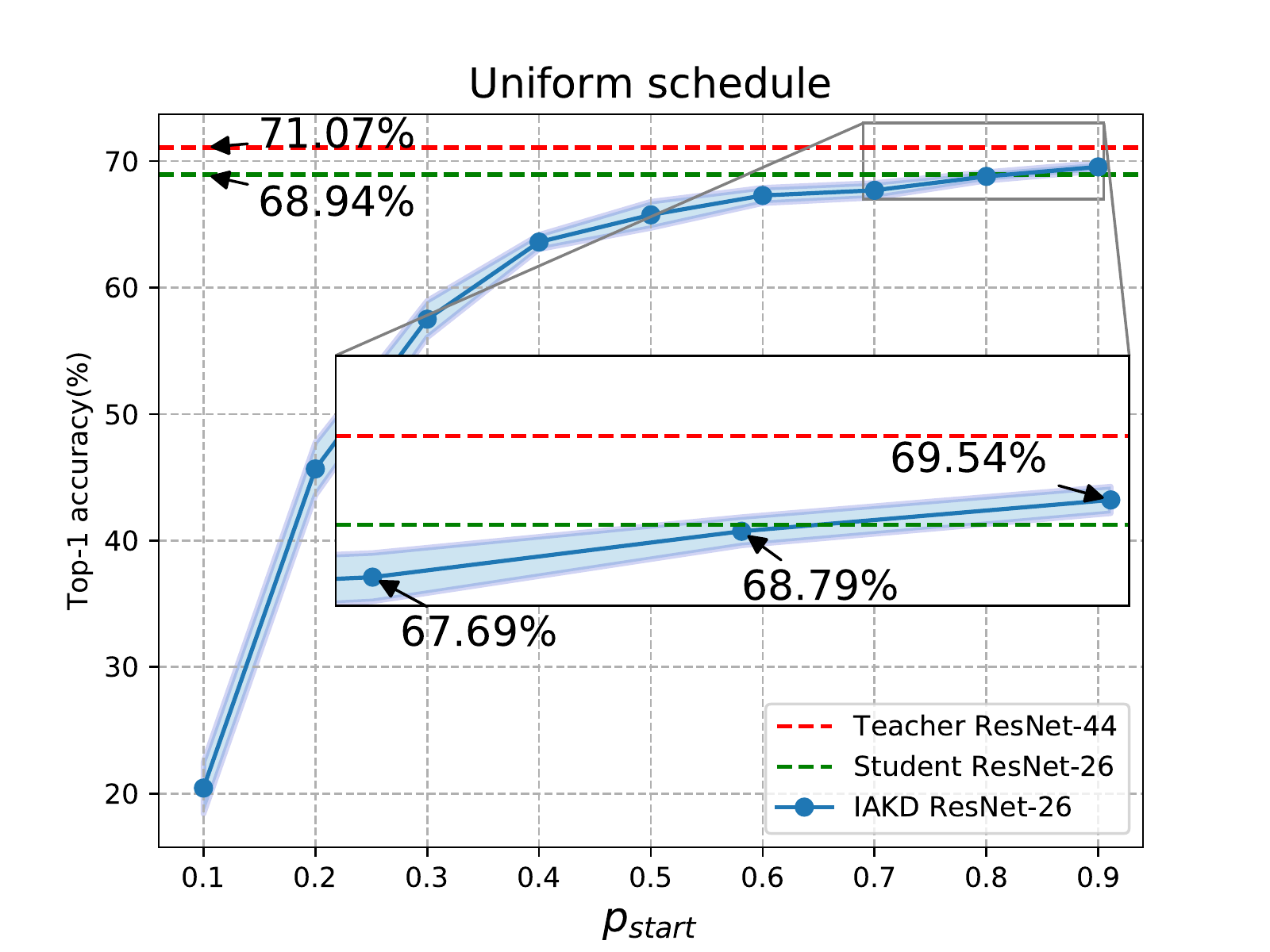}
         \caption{CIFAR-100}
         \label{fig:cifar100_uniform_results}
     \end{subfigure}
        \caption{The results of IAKD-U with different $p_{start}$ . The figure shows Top-1 accuracy on test datasets. Each experiment is repeated 5 times, and we take the mean value as the result.}
        \label{fig:experimental results with uniform schedule}
\end{figure}

The experimental results are shown in Fig.~\ref{fig:experimental results with uniform schedule}. 
Note that $p_{start}=0$ and $p_{start}=1$ are not reasonable options. When $p_{start}=0$, we train the teacher network with many frozen pretrained blocks.
When $p_{start}=1$, we just train the student network individually without interaction.
As we can see from Fig.~\ref{fig:experimental results with uniform schedule}, IAKD with the uniform schedule (denoted as IAKD-U for short) can actually improve the performance of the student network when the value of $p_{start}$ is large (CIFAR-10: $p_{start}\in\{0.7, 0.8, 0.9\}$, CIFAR-100: $p_{start}=0.9$). 
However, we notice that the performance of the student drops dramatically when the value of $p_{start}$ is small (for example, $p_{start}=0.1$ for both CIFAR-10 and CIFAR-100). This is due to the small value of $p_{start}$ results in frequently choosing the teacher path in each hybrid block, which leads to the insufficient training of the student network.

\subsubsection{Linear Growth Schedule}
We still employ ResNet-44/ResNet-26 as the T/S pair when conducting experiments with the linear growth schedule. 
In this schedule, $p_{i}$ increases linearly from $p_{start}$  for the first epoch to 1 for the last epoch.
During early epochs when the value of $p_{i}$ is relatively small, the selected student blocks and the shared parts frequently interact with the frozen teacher blocks, which means those blocks are supposed to have better initialization when the value of $p_{i}$ is relatively large in late epochs (when the hybrid network is more ``student-like"). 
As $p_{i}$ gradually increases, the hybrid network transitions from ``teacher-like" to ``student-like".  Such distillation process is related to the two-stage knowledge distillation methods~\cite{romero2014fitnets,yim2017gift, ABdistill}.
In these methods, the student network is better initialized based on the distillation losses in the first stage, and is trained for the main task based on the task-specific loss in the second stage.
However, the two-stage knowledge distillation methods make the abrupt transition from the initialization stage to the task-specific training stage. 
As a result, the teacher's knowledge may disappear as the training process proceeds, because the teacher is not involved in guiding the student in the second stage. On the contrary, as $p_{i}$ increases linearly, IAKD makes a smooth transition to the task-specific training.

\begin{figure}
     \centering
     \begin{subfigure}[b]{0.24\textwidth}
         \centering
         \includegraphics[width=\textwidth]{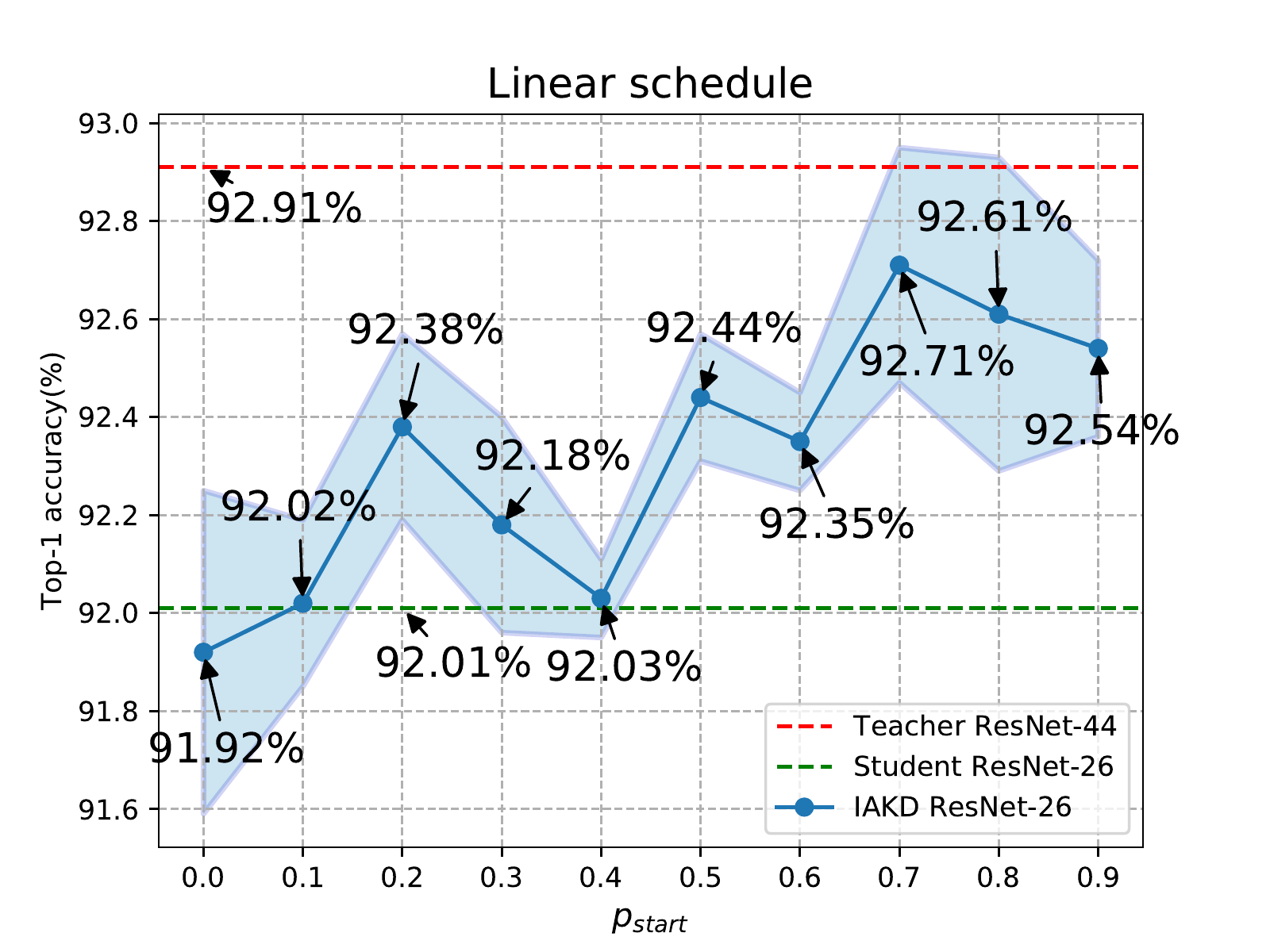}
         \caption{CIFAR-10}
         \label{fig:cifar10_linear_growth_results}
     \end{subfigure}  
     \begin{subfigure}[b]{0.24\textwidth}
         \centering
         \includegraphics[width=\textwidth]{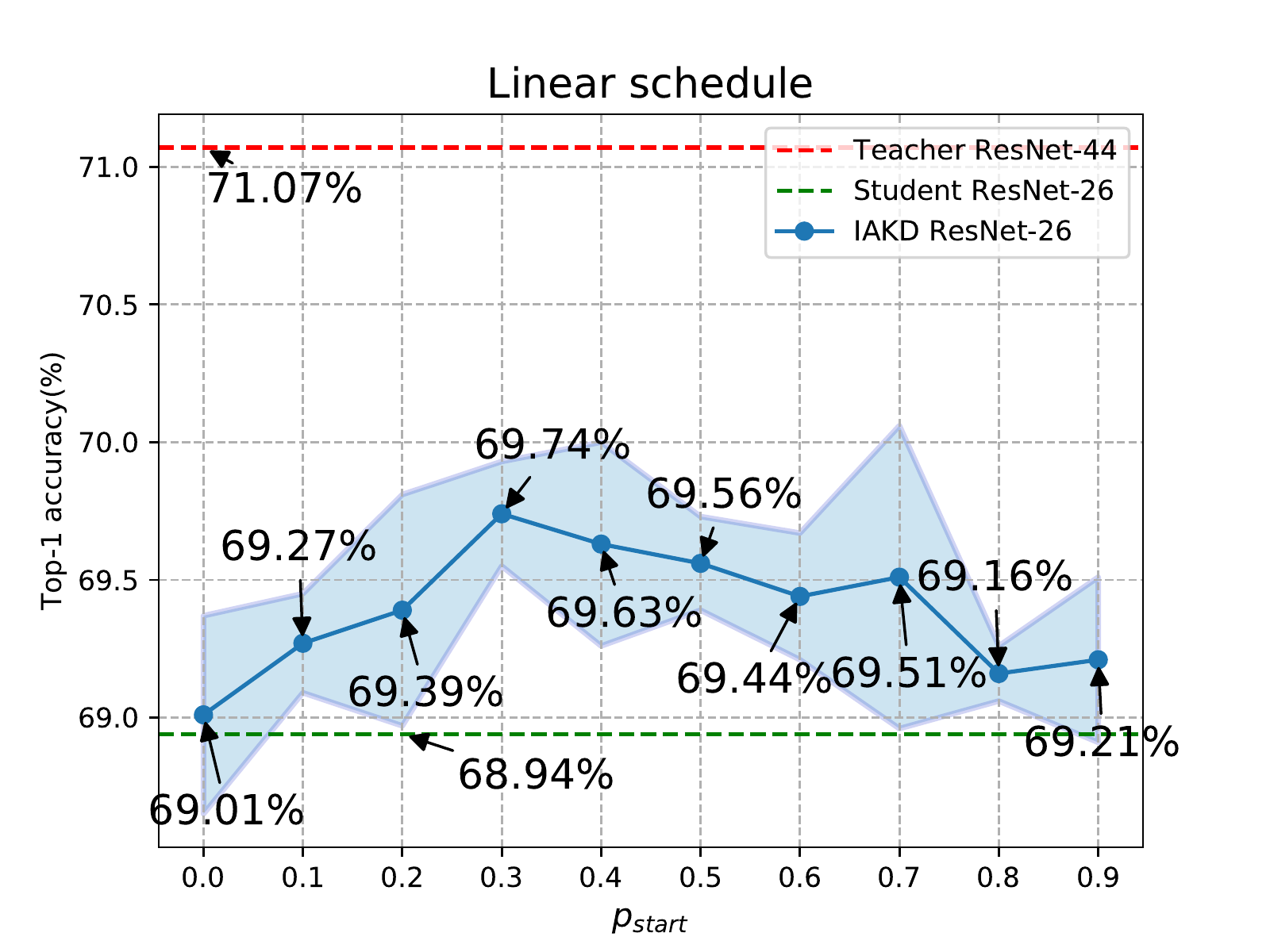}
         \caption{CIFAR-100}
         \label{fig:cifar100_linear_growth_results}
     \end{subfigure}
        \caption{The results of IAKD-L with different $p_{start}$ . The figure shows Top-1 accuracy on test datasets. Each experiment is repeated 5 times, and we take the mean value as the result.}
        \label{fig:experimental results with linear schedule}
\end{figure}

The experimental results of IAKD with linear growth schedule (denoted as IAKD-L) are shown in Fig.~\ref{fig:experimental results with linear schedule}. 
Comparing Fig.~\ref{fig:experimental results with linear schedule} with Fig.~\ref{fig:experimental results with uniform schedule}, we have two main observations.
The first is IAKD-L improves the original student performance on almost every value of $p_{start}$ while IAKD-U only works for large $p_{start}$. 
The second is the best distillation results of IAKD-L on both CIFAR-10 and CIFAR-100 datasets are better than the ones of IAKD-U. 
These two observations indicate that, as the student network becomes powerful, the interaction is more effective between the teacher network and student network when at a gradually reduced rate other than a fixed level. 

\subsubsection{Review Schedule}
\label{sec:review schedule}
\begin{figure}
     \centering
     \begin{subfigure}[b]{0.24\textwidth}
         \centering
         \includegraphics[width=\textwidth]{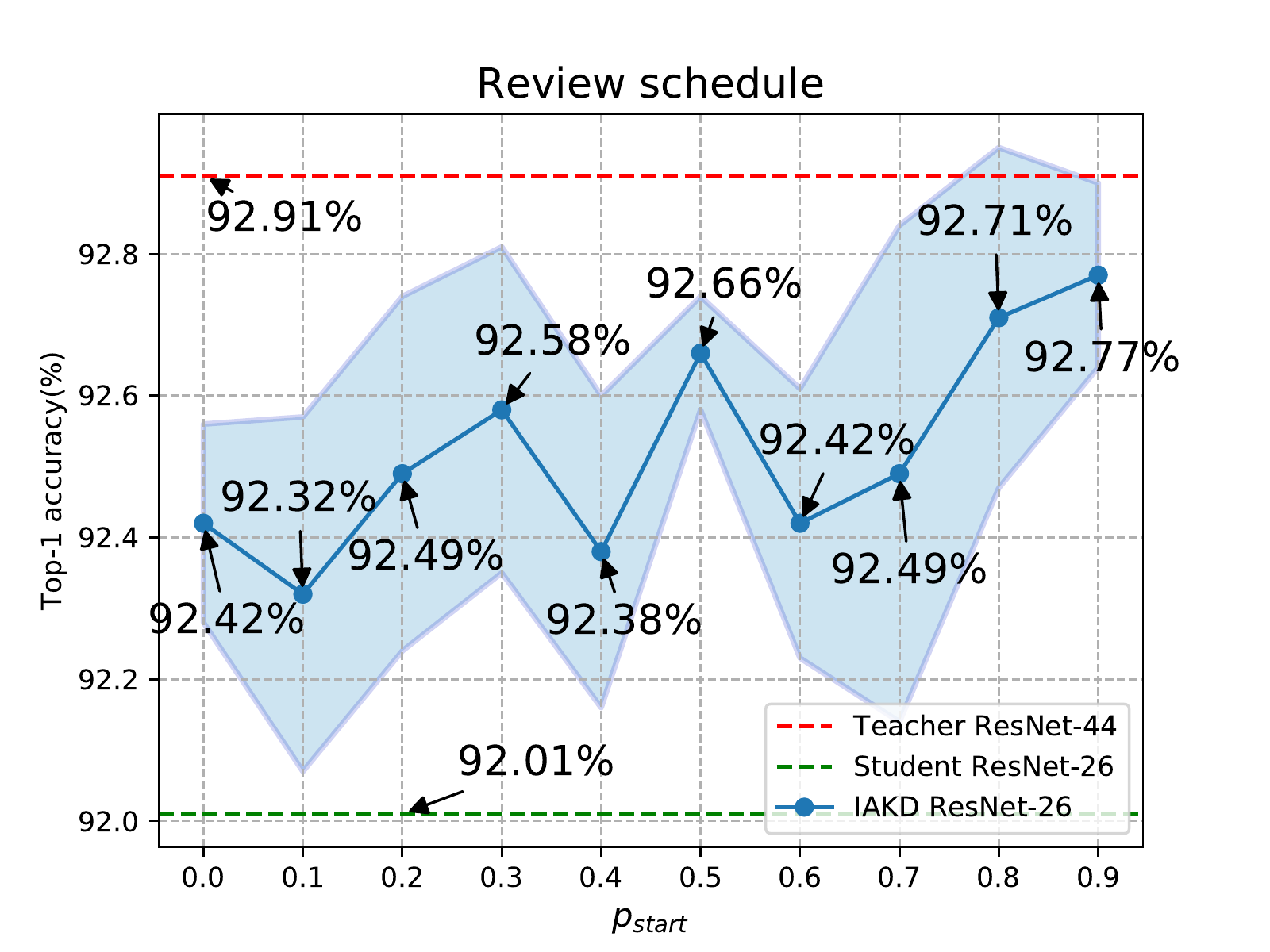}
         \caption{CIFAR-10}
         \label{fig:cifar10_review_results}
     \end{subfigure}  
     \begin{subfigure}[b]{0.24\textwidth}
         \centering
         \includegraphics[width=\textwidth]{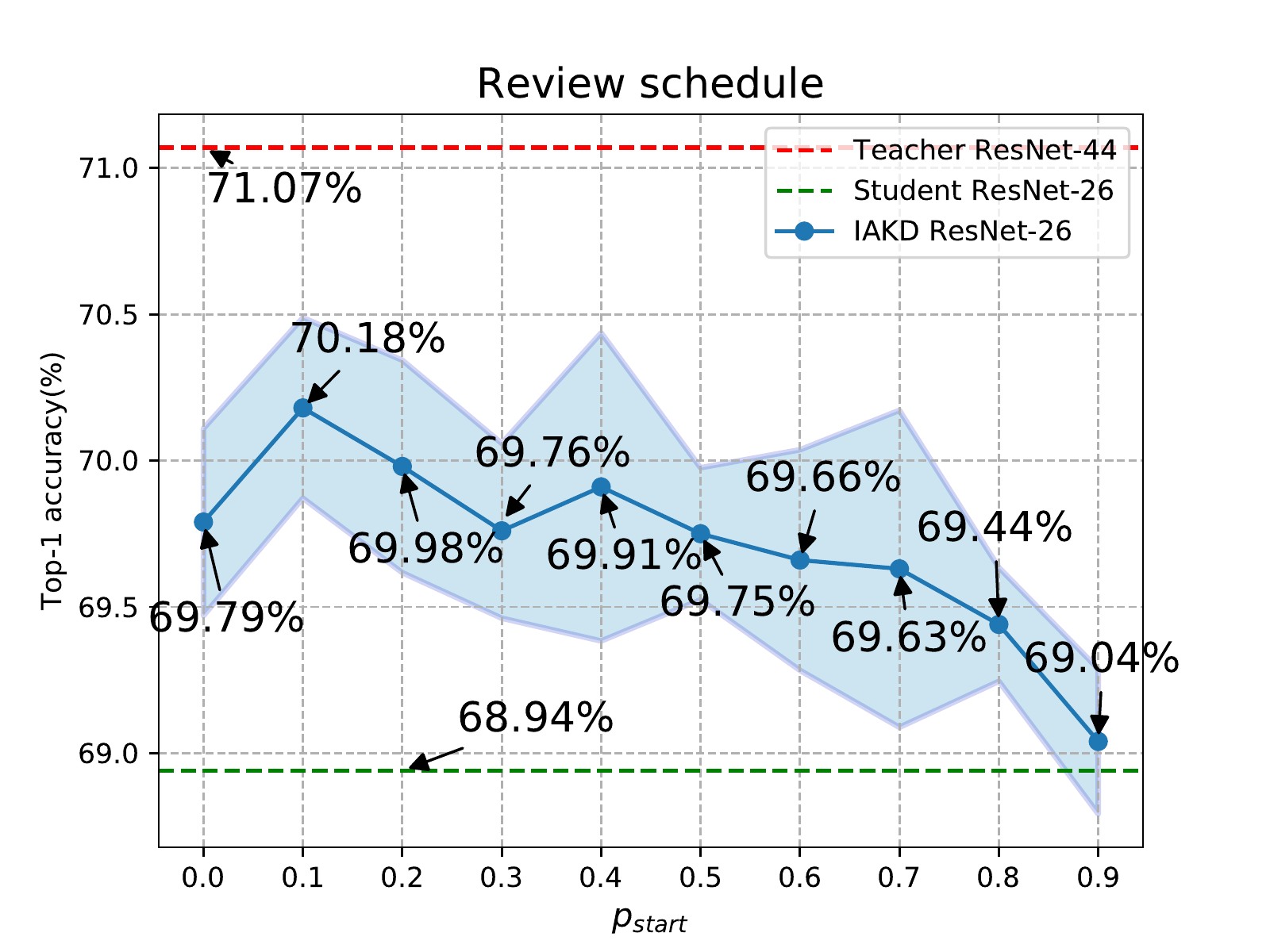}
         \caption{CIFAR-100}
         \label{fig:cifar100_review_results}
     \end{subfigure}
      \caption{The results of IAKD-R with different $p_{start}$ . The figure shows Top-1 accuracy on test datasets. Each experiment is repeated 5 times, and we take the mean value as the result.}
        \label{fig:experimental results with review schedule}
\end{figure}

For fair comparison, we employ ResNet-44/ResNet-26 as the T/S pair to conduct experiments on IAKD with the review schedule (denoted as IAKD-R for short). 
The experimental results of IAKD-R are shown in Fig.~\ref{fig:experimental results with review schedule}. IAKD-R, which considers the change of learning rate, is actually an advanced version of IAKD-L.
Comparing Fig.~\ref{fig:experimental results with review schedule} with Fig.~\ref{fig:experimental results with linear schedule}, IAKD-R could further improve the performance of the original student in contrast to IAKD-L. 
Even for very small value of $p_{start}$, IAKD-R still provides improvement for the original student network. 

\begin{table}[t]
\renewcommand\arraystretch{1.2}
\centering
\caption{The best results of each probability schedule on CIFAR-10 and CIFAR-100. The numbers in brackets represent the value of $p_{start}$ when achieving the corresponding result. The best performance among different probability schedules for each dataset is shown in \textbf{bold}.}
\begin{tabular}{|c|c|c|}
\hline
Method/Model & CIFAR-10               & CIFAR-100              \\ \hline 
ResNet-44 (Teacher)         & 92.91\%                & 71.07\%                \\ \hline
ResNet-26 (Student)         & 92.01\%                & 68.94\%                \\ \hline \hline
IAKD-U                      & 92.55\%$\pm$0.11\% (0.9)          & 69.54\%$\pm$0.32\% (0.9)          \\ \hline
IAKD-L                     & 92.71\%$\pm$0.24\% (0.7)          & 69.74\%$\pm$0.19\% (0.3)          \\ \hline
IAKD-R                      & \textbf{92.77\%$\pm$0.13\% (0.9)} & \textbf{70.18\%$\pm$0.31\% (0.1)} \\ \hline
\end{tabular}
\label{tab:different kind of schedules}
\end{table}
For more clearly comparison, we summarize the best results of different probability schedules in Tab.~\ref{tab:different kind of schedules}. 
It can be seen that IAKD-R obtains the highest classification accuracy in comparison with IAKD-U and IAKD-L on both CIFAR-10 and CIFAR-100 datasets. 
From above discussions and comparisons, we can conclude that compared with uniform schedule and linear growth schedule, review schedule is the most effective interaction mechanism which makes the teacher network re-interact with the student network once the learning rate decreases.
Thus, we take the review schedule as our final probability schedule setting when comparing with other state-of-the-art knowledge distillation methods.
Besides, it is worth noticing that for IAKD-L and IAKD-R, we need to set $p_{start}$ to a small value when the number of classes is large, while we need to set $p_{start}$ to a large value when the number of classes is small.
This reflects that the distillation process can be assisted well by adequate interaction based on the difficulty of a certain task.

Moreover, to ensure the generalization of the proposed review schedule, we conduct more experiments with different schedules, using different architecture combinations. The value of $p_{start}$ in each probability schedule on CIFAR-100 is the same as the one in Tab.~\ref{tab:different kind of schedules}. Experimental results are shown in Tab.~\ref{tab:diff_arch_diff_schedule}. As we can see, review schedule (\ie IAKD-R) consistently outperforms other two schedules  (IAKD-U and IAKD-L) and can achieve promising performance improvement over the student networks in different teacher/student settings, proving the generalization of the review schedule.
\begin{table}[t]
\renewcommand\arraystretch{1.2}
\centering
\caption{Evaluations on different architecture combinations with different probability schedules. Average over 5 runs on CIFAR-100. The table gives the Top-1 accuracy. The best results are shown in \textbf{bold}.}
\resizebox{0.48\textwidth}{!}{
\begin{tabular}{|c|c|c|c|}
\hline
Setup & (a) & (b) & (c) \\
Teacher     & ResNet-18~\cite{he2016deep}  & VGG-19~\cite{simonyan2014very} & ResNet-50x4~\cite{he2016deep}    \\ 
Student      & VGG-8~\cite{simonyan2014very} & VGG-11~\cite{simonyan2014very}  & ShuffleNet-v1~\cite{zhang2018shufflenet} \\ \hline \hline
Teacher            & 72.32\% & 72.74\% & 72.79\% \\ \hline
Student           & 68.91\% & 68.84\% & 67.11\% \\ \hline

IAKD-U ($p_{start}$=0.9)   & 69.60\%($\pm$0.12\%) & 70.21\%($\pm$0.27\%) & 66.13\%($\pm$0.41\%)\\ \hline

IAKD-L ($p_{start}$=0.3)  & 68.98\%($\pm$0.20\%) & 69.96\%($\pm$0.30\%) &66.47\%($\pm$0.27\%) \\ \hline

IAKD-R ($p_{start}$=0.1) & \textbf{69.80\%($\pm$0.23\%)} & \textbf{70.94\%($\pm$0.15\%)} & \textbf{68.25\%($\pm$0.10\%)} \\ \hline

\end{tabular}}
\label{tab:diff_arch_diff_schedule} 
\end{table}

\subsection{Study of Interaction Mechanism}
In this subsection, the training settings are the same with those in Sec.~\ref{sec.4.1}. 
Each experiment is repeated 5 times, and we take the mean value as the final result.
\subsubsection{Without Interaction}
By randomly swapping in the teacher blocks at each iteration, we achieve the interaction between the teacher and the student. 
To investigate the effectiveness of the interaction mechanism, we turn off the interaction between the teacher network and the student one during the distillation process, \ie we do not replace the student block with the corresponding teacher blocks.
Correspondingly, taking the whole pretrained teacher network and the student network, we force the output of each student block to be similar with the output of the corresponding teacher blocks.
%
%
Given that IAKD randomly swaps in the teacher blocks to replace the student block, we can randomly measure the difference between the output of each student block and that of the corresponding teacher blocks.
The student network is trained by the following loss function:
\begin{equation}
	\label{eq:no_swap_in}
    \mathcal{L}_{NIA} =\gamma(y,\hat{y}_{s})+ \sum_{n \in B} a_{n}\varphi (F_{n}^{S}, F_{n}^{T}),
\end{equation}
where $y$ is the ground truth. $\hat{y}_{s}$ is the output of the student network. $B$ is the number of student blocks except the blocks shared by the teacher and the student. $F_{n}^{S}$ and $F_{n}^{T}$ are the outputs of the student block and the corresponding teacher blocks, respectively. NIA denotes No Interaction between the student and the teacher.
$\gamma(\cdot,\cdot)$ is the cross-entropy loss. $\varphi (\cdot, \cdot)$ is the $L_{2}$ loss, penalizing the difference between $F_{n}^{S}$ and $F_{n}^{T}$. $a_{n} \in \{0,1\}$ denotes a Bernoulli random variable, and indicates whether the $n$-th loss is added ($a_{n}=1$) or ignored ($a_{n}=0$).
We denote $p_{n}=P(a_{n}=0)$, which represents the probability of ignoring the $n$-th loss. For simplicity and fair comparison, we set the probability schedule of $p_{n}$ to be the same as the probability schedule of $p_{i}$ in IAKD-U.



The experimental results are shown in Tab.~\ref{tab:cifar10} and Tab.~\ref{tab:cifar100}. 
Comparing with IAKD-U when showing better performance improvement on CIFAR-10 ($p_{i}\in\{0.7, 0.8, 0.9\}$) and CIFAR-100 ($p_{i}=0.9$), the student networks trained without interaction (w/o IA) achieve slight performance improvement. 
This proves that interaction mechanism well assists the student to achieve much better performance. 
\subsubsection{Group-wise Interaction}
Until now, we achieve IAKD by block-wise interaction (one student block pairs with several teacher blocks). 
To verify the effectiveness of block-wise interaction, we reduce the interaction level to group-wise~\cite{Zagoruyko2017AT} interaction (one student group pairs with one teacher group). 
For simplicity, we only choose IAKD-U to achieve group-wise interaction (denoted as IAKD-U-G). 
ResNet-44/ResNet-26 are employed as the T/S pair. 
As we can see from Tab.~\ref{tab:cifar10} and Tab.~\ref{tab:cifar100}, IAKD-U outperforms IAKD-U-G on both datasets.
Besides, IAKD-U-G fails to achieve performance improvement on CIFAR-100, because of the significant decrease in the interaction level caused by the group-wise interaction.
The student blocks, except the first or the last one in each group, have no chance to interact with the teacher blocks.


\begin{table}[t]
\renewcommand\arraystretch{1.2}
\centering
\caption{Experimental results on CIFAR-10 when training the student without interaction (denoted as ``w/o IA") or with group-wise interaction (denoted as ``IAKD-U-G"). Better performance is shown in \textbf{bold}.}
\resizebox{0.48\textwidth}{!}{
\begin{tabular}{|c|c|c|c|c|}
\hline
Model               & w/o IA & IAKD-U-G & IAKD-U   & $p_{n}$ \\ \hline
ResNet-44 (Teacher) & \multicolumn{3}{c|}{92.91\%} & -    \\ \hline
ResNet-26 (Student) & \multicolumn{3}{c|}{92.01\%} & -    \\ \hline \hline
ResNet-26           & 92.10\%($\pm$0.02\%) &    92.05\%($\pm$0.21\%)     & \textbf{92.11\%($\pm$0.41\%)}&0.7  \\ \hline
ResNet-26           & 92.18\%($\pm$0.05\%) &    92.28\%($\pm$0.05\%)      &\textbf{92.31\%($\pm$0.30\%)}&0.8  \\ \hline
ResNet-26           & 92.18\%($\pm$0.06\%) &    92.37\%($\pm$0.05\%)      & \textbf{92.55\%($\pm$0.11\%)}&0.9  \\\hline 
\end{tabular}
}
\label{tab:cifar10}
\end{table}

\begin{table}[t]
\renewcommand\arraystretch{1.2}
\centering
\caption{Experimental results on CIFAR-100 when training the student without interaction (denoted as ``w/o IA") or with group-wise interaction (denoted as ``IAKD-U-G"). Better performance is shown in \textbf{bold}.}
\resizebox{0.48\textwidth}{!}{
\begin{tabular}{|c|c|c|c|c|}
\hline
Model               & w/o IA & IAKD-U-G & IAKD-U   & $p_{n}$ \\ \hline 
ResNet-44 (Teacher) &  \multicolumn{3}{c|}{71.07\%} & -    \\ \hline
ResNet-26 (Student) &  \multicolumn{3}{c|}{68.94\%} & -    \\ \hline \hline
ResNet-26           & 69.20\%($\pm$0.15\%) & 68.68\%($\pm$0.16\%) &   \textbf{69.54\%($\pm$0.32\%)} &0.9  \\ \hline
\end{tabular}
}
\label{tab:cifar100}
\end{table}

\subsection{Discussions about More Knowledge Distillation Points}
From the view of the conventional, \textit{non-interactive} knowledge distillation methods, the end of each set of swapped-in teacher blocks can be regarded as the knowledge distillation point where the knowledge of the teacher network is transferred to the student network. Randomly swapping in the teacher blocks to replace the student blocks implicitly increases the number of knowledge distillation points. To verify more knowledge distillation points do not necessarily lead to more performance improvement and may even harm the student performance, we introduce more connections between the teacher and the student to multi-connection knowledge based methods (\ie AT~\cite{Zagoruyko2017AT} and FSP~\cite{yim2017gift}). Accordingly, we modify AT to transfer attention maps at the end of each student block rather than the end of each group, and name it as 9-point AT because there are totally 9 knowledge distillation points. Similarly, we modify FSP to calculate the Gramian matrix by the input and the output of each student block instead of each group, and name it as 9-point FSP. The experiments are conducted by ResNet-44/ResNet-26 pair on CIFAR-10 and CIFAR-100 datasets.

\begin{table}[t]
\renewcommand\arraystretch{1.2}
\centering
\caption{The experimental results when non-interactive methods take more knowledge distillation points. The table shows the Top-1 accuracy on test datasets. Average over 5 runs on CIFAR-100. The best results are shown in \textbf{bold}.}
\begin{tabular}{|c|c|c|}
\hline
Model/Method        & CIFAR-10 & CIFAR-100 \\ \hline \hline
ResNet-44 (Teacher) & 92.91\%  & 71.07\%   \\ \hline
ResNet-26 (Student) & 92.01\%  & 68.94\%   \\ \hline
9-point AT          & 92.01\%($\pm$0.19\%)  & 69.56\%($\pm$0.22\%)   \\ \hline
AT                  & 92.55\%($\pm$0.12\%)  & 69.72\%($\pm$0.17\%)   \\ \hline
9-point FSP         & 92.45\%($\pm$0.27\%)  & 69.07\%($\pm$0.43\%)   \\ \hline
FSP                 & 92.47\%($\pm$0.06\%)  & 69.10\%($\pm$0.34\%)   \\ \hline
IAKD-R (Ours)         & \textbf{92.77\%($\pm$0.13\%)}  & \textbf{70.18\%($\pm$0.31\%)}   \\ \hline
\end{tabular}
\medskip
\label{tab:more knowledge distillation points}
\end{table}

Experimental results are shown in Tab~\ref{tab:more knowledge distillation points}. It turns out, for conventional, \textit{non-interactive} knowledge distillation methods, more knowledge distillation points do not necessarily lead to more performance improvement. For AT, more distillation points even make the distillation performance worse than the performance of the original AT. This is because excessive supervision causes a hard constraint for the student network. However, as shown in Tab.~\ref{tab:different kind of schedules} and Tab.~\ref{tab:diff_arch_diff_schedule}, introducing more possible distillation points even improves the distillation performance of our proposed IAKD. 
\subsection{Comparison with State-of-the-arts}
\label{sec:4.3}
We compare our proposed IAKD-R with other representative state-of-the-art knowledge distillation methods: HKD~\cite{hinton2015distilling}, AT~\cite{Zagoruyko2017AT}, FSP~\cite{yim2017gift}, Feature distillation~\cite{heo2019overhaul}, AB~\cite{ABdistill}, VID-I~\cite{ahn2019variational} and RKD-DA~\cite{park2019relational}.
As the discussion above, all these comparison methods are  \textit{non-interactive}. 
We conduct experiments on three datasets: CIFAR-10~\cite{krizhevsky2009learning}, CIFAR-100~\cite{krizhevsky2009learning}, and TinyImageNet~\cite{yao2015tiny}. TinyImageNet contains images of size $64\times64$ with 200 classes. Each experiment is repeated 5 times, and we take the mean value as the final report result. 
For other compared methods, we directly use the author-provided codes if publicly available, or implement those methods based on the original paper. The hyper-parameters of contrastive methods are listed as follows.
\begin{enumerate}
\item  \textbf{HKD}~\cite{hinton2015distilling}: temperature $\mathcal{T}$=$4$, balance factor: $\alpha$=0.9 (Eq.~20 in \cite{tian2019crd}).
\item  \textbf{AT}~\cite{Zagoruyko2017AT}: balance factor: $\beta$=1000 (Eq.~2 in \cite{Zagoruyko2017AT}).
\item  \textbf{FSP}~\cite{yim2017gift}: No balance factor for the task-specific and the distillation loss because the distillation is in the initialization stage where only the distillation loss applies. In the initialization stage, we set $\lambda_{i}$, the hyper-parameter to balance each FSP loss term in Eq.~2 in \cite{yim2017gift}, to 1.
\item  \textbf{AB}~\cite{ABdistill}: No balance factor for the task-specific loss and the distillation loss because the distillation is in the initialization stage where only the distillation loss applies. In the initialization stage, there are three hyper-parameters: $\lambda_1$, $\lambda_2$ and $\lambda_3$ to balance the different terms of the distillation loss. We set $\lambda_1$, $\lambda_2$ and $\lambda_3$ to $\frac{1}{4}\times10^{-3}$, $\frac{1}{2}\times10^{-3}$, and $1\times10^{-3}$, respectively (See \cite{ABdistill} for explanations of these hyper-parameters).
\item  \textbf{VID-I}~\cite{ahn2019variational}: balance factor: $\lambda_{k}$=1 (Eq.~2 in \cite{ahn2019variational}).
\item  \textbf{RKD-DA}~\cite{park2019relational}: balance factors: $\lambda_{RKD-D}$=25 and $\lambda_{RKD-A}$=50 (See \cite{park2019relational} for explanations of these hyper-parameters).
\item  \textbf{Feature distillation}~\cite{heo2019overhaul}: balance factor: $\alpha$=$10^{-3}$ (Eq.~6 in \cite{heo2019overhaul}). In the distillation loss function in Eq. 5 in \cite{heo2019overhaul}, the loss is multiplied by 2 when the spatial size of the feature maps is decreased by twice as the input moves from shallow to deep layers.
\end{enumerate}

For fair comparison, the optimization configurations are identical for all methods on one dataset.

\begin{table*}[t]
\caption{Comparison of test accuracy by different knowledge distillation methods on CIFAR-10, CIFAR-100, and TinyImageNet datasets.\ The best and second best results on Top-1 accuracy are shown in \textbf{bold} and \underline{underlined}, respectively.}
\renewcommand\arraystretch{1.4}
\centering
\resizebox{0.9\textwidth}{!}{
\begin{tabular}{|c|c|c|c|c|c|c|c|} 
\hline
\multirow{2}{*}{\textbf{Method} } 
    & \multirow{2}{*}{\textbf{Model} } 
        & \multicolumn{2}{c|}{\textbf{CIFAR-10} } 
            \
                & \multicolumn{2}{c|}{\textbf{CIFAR-100} } 
                    \
                        & \multicolumn{2}{c|}{\textbf{TinyImageNet}}
                            \
\\ 
\cline{3-8} 
\   
    &  \                                
        & Size/FLOPs
            & Top-1             
                & Size/FLOPs
                    & Top-1             
                        & Size/FLOPs                      
                            & Top-1             
\\ 
\hline\hline
Teacher                           
    & ResNet-44/80$^{*}$               
        & 0.66M/98.51M
            & 92.91\%
                & 0.67M/98.52M
                    & 71.07\%
                        & 1.26M/737.24M
                            & 55.73\%
\\ 
\hline
Student 
    & \multirow{10}{*}{ResNet-26} 
        & \multirow{10}{*}{0.37M/55.62M} 
            & 92.01\%
                & \multirow{10}{*}{0.38M/55.62M} & 68.94\% 
                    & \multirow{10}{*}{0.38M/222.47M} & 50.04\%
\\ 
\cline{1-1}\cline{4-4}\cline{6-6}\cline{8-8}
HKD~\cite{hinton2015distilling}                               
    & 
        &  \
            & \underline{92.90\%($\pm$0.19\%)}  
                &  \
                    & 69.59\% ($\pm$0.26\%) 
                        &  \
                            & 50.25\%($\pm$0.46\%)
\\ 
\cline{1-1}\cline{4-4}\cline{6-6}\cline{8-8}
AT~\cite{Zagoruyko2017AT}                               
    & 
        &  \
            & 92.55\%($\pm$0.12\%) 
                &  \
                    & 69.72\%($\pm$0.17\%)
                        &  \
                            & 52.07\%($\pm$0.21\%)
\\ 
\cline{1-1}\cline{4-4}\cline{6-6}\cline{8-8}
FSP~\cite{yim2017gift}                      
    &                      
        &  \
            & 92.47\%($\pm$0.06\%)
                &  \    
                    & 69.10\%($\pm$0.34\%)
                        &  \    
                            & 50.97\%($\pm$0.50\%)
\\ 
\cline{1-1}\cline{4-4}\cline{6-6}\cline{8-8}
AB~\cite{ABdistill}         
    & 
        &  \
            & 92.59\%($\pm$0.16\%)
                &  \
                    & 69.40\%($\pm$0.42\%) 
                        &  \            
                            & 48.48\%($\pm$2.96\%)
\\ 
\cline{1-1}\cline{4-4}\cline{6-6}\cline{8-8}
VID-I~\cite{ahn2019variational}       
    & 
        &  \
            & 92.48\%($\pm$0.10\%)
                &  \
                    & 69.90\%($\pm$0.23\%)
                        &  \
                            & 51.98\%($\pm$0.15\%)
\\ 
\cline{1-1}\cline{4-4}\cline{6-6}\cline{8-8}
RKD-DA~\cite{park2019relational}    
    &  
        &  \
            & 92.70\%($\pm$0.16\%) 
                &  \
                    & 69.46\%($\pm$0.39\%)  
                        &  \
                            & 51.69\%($\pm$0.43\%)
\\ 
\cline{1-1}\cline{4-4}\cline{6-6}\cline{8-8}
Feature distillation~\cite{heo2019overhaul}                         
    &  
        & \
            & 92.29\%($\pm$0.10\%) 
                & \
                    & 69.11\%($\pm$0.17\%)
                        & \                     
                            & 52.56\%($\pm$0.22\%) 
\\ 
\cline{1-1}\cline{4-4}\cline{6-6}\cline{8-8}
IAKD-R (Ours)  
    &  
        & \
            & 92.77\%($\pm$0.13\%)
                & \
                    & \underline{70.18\%($\pm$0.31\%)}  
                        & \
                            & \underline{53.28\%($\pm$0.13\%)}
\\ 
\cline{1-1}\cline{4-4}\cline{6-6}\cline{8-8}
IAKD-R+HKD (Ours)              
    & 
        & \
            & \textbf{93.31\%($\pm$0.08\%)}  
                & \
                    & \textbf{71.02\%($\pm$0.16\%)}  
                        & \ 
                            & \textbf{53.50\%($\pm$0.31\%)}  
\\
\hline
\end{tabular}
}
\begin{tablenotes}
    \item $^{*}$\small{ResNet-44 is used as the teacher on CIFAR-10 and CIFAR-100, while ResNet-80 is used as the teacher on TinyImageNet.}
\end{tablenotes}
\label{tab:comparison}
\end{table*}

For CIFAR-10 and CIFAR-100, we use the same optimization configurations as described in Sec.~\ref{sec.4.1}.
Specially, for two-stage methods like FSP and AB, 50 epochs are used for initialization and 150 epochs are used for classification training. 
The learning rate at the initialization stage is 0.1 and 0.001 for AB and FSP, respectively. 
For AB and FSP, the initial learning rate at the classification training stage is 0.1, and is multiplied by 0.1 at 75, 110 epochs. 
For TinyImageNet, we employ ResNet-80/ResNet-26 as the T/S pair. Horizontal flipping is applied for data augmentation. 
We optimize the network using SGD with batchsize 128, momentum 0.9 and weight decay $1\times10^{-4}$. 
The number of total training epochs is 300. 
The initial learning rate is 0.1 and is multiplied by 0.2 at 60, 120, 160, 200, 250 epochs. 
Specially, for FSP and AB, 50 epochs are used for initialization and 250 epochs are used for classification training. 
The learning rate at the initialization stage is 0.1 and 0.001 for AB and FSP, respectively.  
For AB and FSP, the initial learning rate at the classification training is 0.1, and is multiplied by 0.2 at 60, 120, 160, 200 epochs. 

For the proposed IAKD, we use the review schedule as the final probability schedule setting. $p_{start}$ is set to 0.9 for comparison on CIFAR-10, and is set to 0.1 for comparison on CIFAR-100 and TinyImageNet. 
The configurations are based on the experimental results and arguments in Sec.~\ref{sec:review schedule}. 

In Tab.~\ref{tab:comparison}, we list the comparison results on CIFAR-10, CIFAR-100, and TinyImageNet by different methods.\ 
As we can see, the proposed IAKD improves the performance of the student network more effectively than most contrastive methods.\
Meanwhile, the distillation result of IAKD is competitive with HKD on CIFAR-10.\ We then combine HKD with our proposed IAKD through adding distillation loss to Eq.~\ref{eq:hybrid_loss}, and denote it as IAKD-R+HKD.
Surprisingly, IAKD-R+HKD even outperforms the performance of the original teacher on CIFAR-10.
On CIFAR-100 and TinyImageNet, IAKD-R+HKD outperforms all other contrastive methods.
IAKD-R also shows highly competitive results on CIFAR-100 and TinyImageNet. 
In addition, our proposed method, robust as we can see, consistently achieves significant performance improvement on three datasets while most other contrastive methods do not.
%

\noindent
\textbf{Advantages on learning efficiency}.\
Our proposed IAKD-R is also benefited from better learning efficiency.\ 
Conventional knowledge distillation methods update all student blocks along the whole training iterations.\ 
Our work reveals that this laborious distillation is not necessary for promising performance.\ 
In our IAKD-R, the student blocks replaced by the teacher ones will not be updated at that iteration.\ 
Thus, in each training iteration, a \textit{portion} of the student network is updated.\
For the possible replaced student block, our training epochs number, the mathematical expectation, is 190, 110 and 165 on CIFAR-10, CIFAR-100 and TinyImageNet, respectively, which accounts for 95\%, 55\% and 55\% of other contrastive methods on corresponding datasets. \ 
This demonstrates that our IAKD-R enjoys efficient learning capacity, and also validates that the student network indeed benefits well from the interaction with the teacher network.

As mentioned in Sec.~\ref{sec:intro}, our proposed IAKD discards the cumbersome knowledge transformation process and gets rid of the feature extraction of the entire teacher network, we compare the training time of different knowledge distillation methods, using ResNet-80/ResNet-26 teacher-student pair on TinyImageNet dataset. We conduct training time comparison on a single NVIDIA RTX 2080Ti GPU with Intel Core i7-8700 CPU and PyTorch framework. As shown in Tab.~\ref{tab:time}, our proposed method gives a highly competitive result in contrast to other methods. It only needs 4 hours and 40 minutes to complete the training process of the student network, which reflects that the high training efficiency of our proposed IAKD. 

\begin{table}[htbp]
\renewcommand\arraystretch{1.2}
\centering
\caption{Training time comparison on TinyImageNet.}
\begin{tabular}{|c|c|} 
\hline
Method               & Time Cost  \\ 
\hline\hline
HKD~\cite{hinton2015distilling}                  & 6h 47m    \\ 
\hline
AT~\cite{Zagoruyko2017AT}                   & 11h 29m   \\ 
\hline
FSP~\cite{yim2017gift}                  & 7h 13m    \\ 
\hline
AB~\cite{ABdistill}                   & 4h 21m    \\ 
\hline
VID-I~\cite{ahn2019variational}                & 13h 44m   \\
\hline
RKD-DA ~\cite{park2019relational}              & 6h 48m    \\ 
\hline
Feature distillation~\cite{heo2019overhaul} & 7h 5m     \\ 
\hline
IAKD-R (Ours)          & 4h 40m    \\
\hline
\end{tabular}
\medskip
\label{tab:time}
\end{table}

\subsection{Different Architecture Combinations}
To ensure the applicability of IAKD, we explore more architecture combinations on CIFAR-100, as shown in Tab.~\ref{tab:diff_arch}. We replace the ``Conv3$\times$3-BN-ReLU" with ``group Conv3$\times$3-BN-ReLU-Conv1$\times$1-BN-ReLU" to achieve the cheap ResNet, as suggested by \cite{crowley2018moonshine}. The experimental settings are the same as those described in Sec.~\ref{sec:4.3}. As we can see from Tab.~\ref{tab:diff_arch}, in all the architecture combinations, IAKD-R and IAKD-R+HKD can consistently achieve performance improvement over the student network. On the other hand, contrastive methods give different knowledge distillation performance in different settings. In the case of (a), only IAKD-R and IAKD-R+HKD can successfully achieve performance improvement over the student cheap ResNet-26. AT and RKD-DA fail to improve the student performance in the case of (b). RKD-DA and AB fail to achieve performance improvement in the case of (d). Therefore, the proposed IAKD, a novel knowledge distillation method without additional distillation losses, can also achieve promising performance improvement in different teacher/student settings. 

\begin{table*}[ht]
\centering
\renewcommand\arraystretch{1.3}
\caption{Evaluations on different architecture combinations. Average over 5 runs on CIFAR-100. The table gives the Top-1 accuracy. The best and second best results on Top-1 accuracy are shown in \textbf{bold} and \underline{underlined}, respectively. $\downarrow$ denotes underperforming the corresponding students trained without knowledge distillation.}
\resizebox{0.8\textwidth}{!}{\begin{tabular}{|c|c|c|c|c|}
\hline
Setup & (a) & (b) & (c) & (d) \\
Teacher     & ResNet-44~\cite{he2016deep} & ResNet-18~\cite{he2016deep}  & VGG-19~\cite{simonyan2014very} & ResNet-50x4~\cite{he2016deep}    \\ 
Student     & cheap ResNet-26 & VGG-8~\cite{simonyan2014very} & VGG-11~\cite{simonyan2014very}  & ShuffleNet-v1~\cite{zhang2018shufflenet} \\ \hline \hline
Teacher          & 71.07\%  & 72.32\% & 72.74\% & 72.79\% \\ \hline
Student          & 65.92\%  & 68.91\% & 68.84\% & 67.11\% \\ \hline
HKD~\cite{hinton2015distilling}  & 65.55\%($\pm$0.40\%) $\downarrow$& \textbf{70.63\%($\pm$0.27\%) }& 71.09\%($\pm$0.13\%) & \underline{68.58\%($\pm$0.43\%)} \\ \hline
AT~\cite{Zagoruyko2017AT}          & 65.90\%($\pm$0.37\%) $\downarrow$ & 68.54\%($\pm$0.40\%) $\downarrow$ & 70.93\%($\pm$0.14\%) & 68.13\%($\pm$0.30\%) \\ \hline
RKD-DA~\cite{park2019relational}         & 65.66\%($\pm$0.21\%) $\downarrow$ & 68.69\%($\pm$0.31\%) $\downarrow$ & 70.84\%($\pm$0.14\%) & 66.69\%($\pm$0.38\%) $\downarrow$ \\ \hline
AB~\cite{ABdistill}         & 65.64\%($\pm$0.76\%) $\downarrow$ & 69.86\%($\pm$0.24\%) & \underline{71.21\%($\pm$0.27\%)} & 66.89\%($\pm$1.46\%) $\downarrow$ \\ \hline
IAKD-R & \textbf{67.15\%($\pm$0.33\%)} & 69.80\%($\pm$0.23\%) & 70.94\%($\pm$0.15\%) & 68.25\%($\pm$0.10\%) \\ \hline
IAKD-R+HKD & \underline{66.11\%($\pm$0.72\%)} & \underline{70.51\%($\pm$0.22\%)} & \textbf{71.76\%($\pm$0.21\%)} & \textbf{68.65\%($\pm$0.29\%)} \\ \hline
\end{tabular}}
\label{tab:diff_arch} 
\end{table*}
\section{Conclusion}
\label{sec:conclusion}
In this paper, we proposed a simple yet effective knowledge distillation method named Interactive Knowledge Distillation (IAKD). 
%
By randomly swapping in the teacher blocks to replace the student blocks in each iteration, we accomplish the interaction between the teacher network and the student one.
To properly utilize the interaction mechanism during distillation process, we proposed three kinds of probability schedules. 
Our IAKD does not need additional distillation losses to drive the distillation process, and is complementary to conventional knowledge distillation methods. 
Extensive experiments demonstrated that the proposed IAKD could boost the performance of the student network on diverse image classification tasks.\ 
Instead of mimicking the teacher's feature representation space, our proposed IAKD aims to directly leverage the teacher's powerful feature transformation ability to motivate the student, providing a new perspective for knowledge distillation.

\bibliographystyle{IEEEtran}
\bibliography{IEEEabrv,ref}

\end{document}